\documentclass[runningheads]{llncs}
\usepackage[T1]{fontenc}
\usepackage{graphicx}
\usepackage{cite}
\usepackage{amsmath,amssymb,amsfonts}
\usepackage{algorithmic}
\usepackage{graphicx}
\usepackage{textcomp}
\usepackage{verbatim}
\usepackage{xcolor}
\usepackage{float}
\begin{document}
\title{Feature Representation Learning for NL2SQL Generation Based on
Coupling and Decoupling}
\titlerunning{NL2SQL Generation Based on
Coupling and Decoupling}
%

\author{Chenduo Hao \and
Xu Zhang \and
Chuanbao Gao \and
Deyu Zhou\thanks{Corresponding author}
}

\authorrunning{Chenduo Hao.}
\institute{School of Computer Science and Engineering, Southeast University, Nanjing, China 
\email{213201447@seu.edu.cn, xuzhang123@seu.edu.cn, 213202938@seu.edu.cn,d.zhou@seu.edu.cn}
}
\maketitle              
\begin{abstract}
The NL2SQL task involves parsing natural language statements into SQL queries. While most state-of-the-art methods treat NL2SQL as a slot-filling task and use feature representation learning techniques, they overlook explicit correlation features between the SELECT and WHERE clauses and implicit correlation features between sub-tasks within a single clause. To address this issue, we propose the Clause Feature Correlation Decoupling and Coupling (CFCDC) model, which uses a feature representation decoupling method to separate the SELECT and WHERE clauses at the parameter level. Next, we introduce a multi-task learning architecture to decouple implicit correlation feature representation between different SQL tasks in a specific clause. Moreover, we present an improved feature representation coupling module to integrate the decoupled tasks in the SELECT and WHERE clauses and predict the final SQL query. Our proposed CFCDC model demonstrates excellent performance on the WikiSQL dataset, with significant improvements in logic precision and execution accuracy. The source code for the model will be publicly available on GitHub\footnote{https://github.com/}.

\end{abstract}
\section{Introduction}
\par NL2SQL aims to automate the process of translating natural language (NL) into Structured Query Language (SQL). Its potential applications include table-based question answering. However, constructing an efficient SQL query requires the incorporation of multiple facets of information, including SELECT and WHERE clauses. Consequently, the challenge lies in developing methods to accurately extract these distinct features from the flexible nature of natural language queries.

\par In previous researches, NL2SQL was approached as a sequence-to-sequence (seq2seq) task, as demonstrated in Seq2SQL\cite{DBLP:journals/corr/abs-1709-00103}, coarse-to-fine decoding\cite{dong2018coarse}, and semantic parsing\cite{sun2018semantic}. However, this seq2seq-based approach did not effectively consider the structural information of the SQL query. As a result, the generated SQL queries were often incapable of performing query operations. Recent research, including SQLNet\cite{DBLP:journals/corr/abs-1711-04436} and TypeSQL\cite{yu2018typesql}, utilized a slot-filling method that involves extracting key features to populate corresponding slots. This method eliminates the need to separately consider the structural information of the SQL query. As a result, NL2SQL is transformed into a simple multi-tasking classification task\cite{hwang2019comprehensive,he2019x,lyu2020hybrid}.

\par
\begin{figure*}
\centering
\includegraphics[width=1.0\textwidth,height=0.2\textwidth]{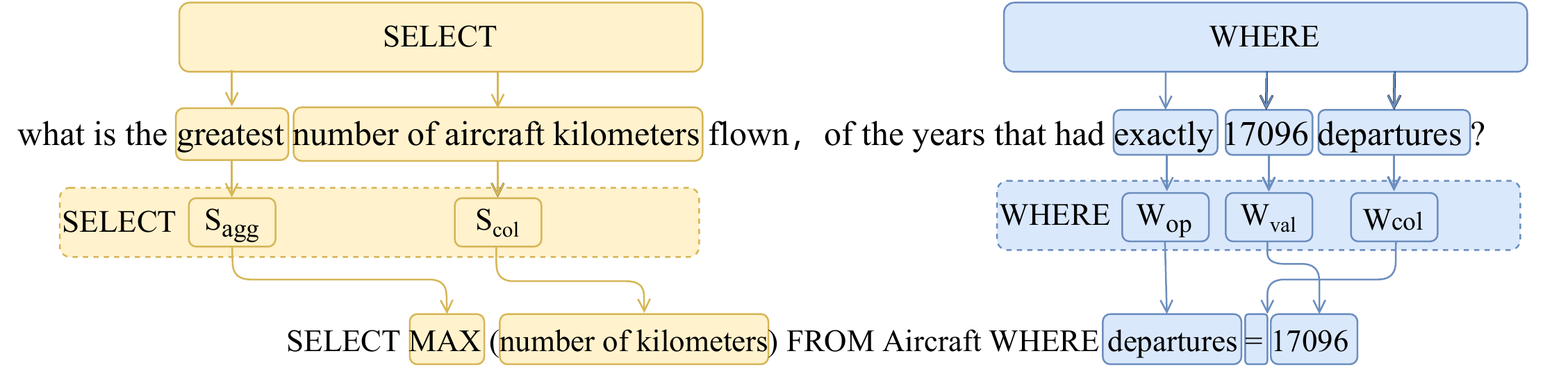}
\caption{
An example of a natural language statement is parsed into a query of SQL language. It can be seen from this instance that the tasks which SELECT and WHERE clauses are responsible for are different, as well as the objects that different tasks are responsible for extracting in their respective modes. It can be seen from the figure that the objects that the SELECT clause and WHERE clauses are responsible for are quite different.} \label{fig1}
\end{figure*}

 However, previous slot-filling-based methods have disregarded two crucial properties \cite{he2019x,lyu2020hybrid}. Firstly, explicit clauses, such as SELECT and WHERE, share dissimilar feature representations, as illustrated in Fig. \ref{fig1}. The SELECT clause is responsible for identifying relevant columns and their operations (e.g., average), while the WHERE clause is responsible for identifying relevant columns, their conditions, and constraints (e.g., operator "="). Moreover, these two parts have different optimization objectives, and they do not influence each other. Consequently, there is a weak correlation between the two explicit clauses. However, previous approaches, such as HydraNet and X-SQL, have employed parameter-sharing layers to train and couple the two clauses, thereby ignoring their weak correlation and leading to reduced model accuracy \cite{torrey2010transfer}. Secondly, implicit correlation features between tasks within the same clause cannot be effectively captured. For instance, X-SQL and HydraNet utilize slot-filling methods to predict SQL components and these methods train different sub-tasks and learn the correlation between them using parameter-sharing layers. However, these approaches overlook the implicit correlation features between different tasks that predict SQL components (e.g., $W_{col}$ (department) and $W_{op}$ (=)). The direct use of parameter sharing may result in the seesaw phenomenon, which exhibits inconsistent performance gains across tasks \cite{tang2020progressive}.
\par In order to address the two aforementioned issues, we present the Clause Feature Correlation Decoupling and Coupling model (CFCDC) that offers two distinct solutions. The first solution involves the hard decoupling of the SELECT and WHERE clauses through a method that employs hard feature representation decoupling. This method decouples the correlation between the SELECT and WHERE clauses from the parameter level and training tasks in each clause separately to achieve the hard decoupling. The second solution is the soft decoupling of the SELECT (or WHERE) clause's implicit correlation feature representation using a composite expert network, which is also leveraged to capture the implicit correlation feature representation between tasks and to mitigate the seesaw phenomenon associated with the use of a multi-task learning approach. Finally, a coupling module based on multi-task learning is utilized to combine the output of our decoupling modules and improve the accuracy of the generated SQL components. CFCDC has demonstrated remarkable performance on the WikiSQL dataset, exhibiting significant improvements in both logic precision and execution precision.

\section{Proposed Method}
This section describes the key techniques employed in our proposed model. To address a specific question \textit{q} along with the corresponding candidate columns \textit{$c_1$,$c_2$,...$c_i$}, a collection of relations is established in the form of the problem and the associated candidate columns. This set of relations is formulated as follows:
\begin{align}
input = contact((c_i,\gamma_i,\delta_i),q_i)
\end{align}
where $\gamma_i$ is the type of $c_i$, $\delta_i$ is the specific name of $c_i$, and contact is a function to concatenate text into a string. 

\begin{figure*}[h]
\centering
\includegraphics[width=0.85\textwidth]{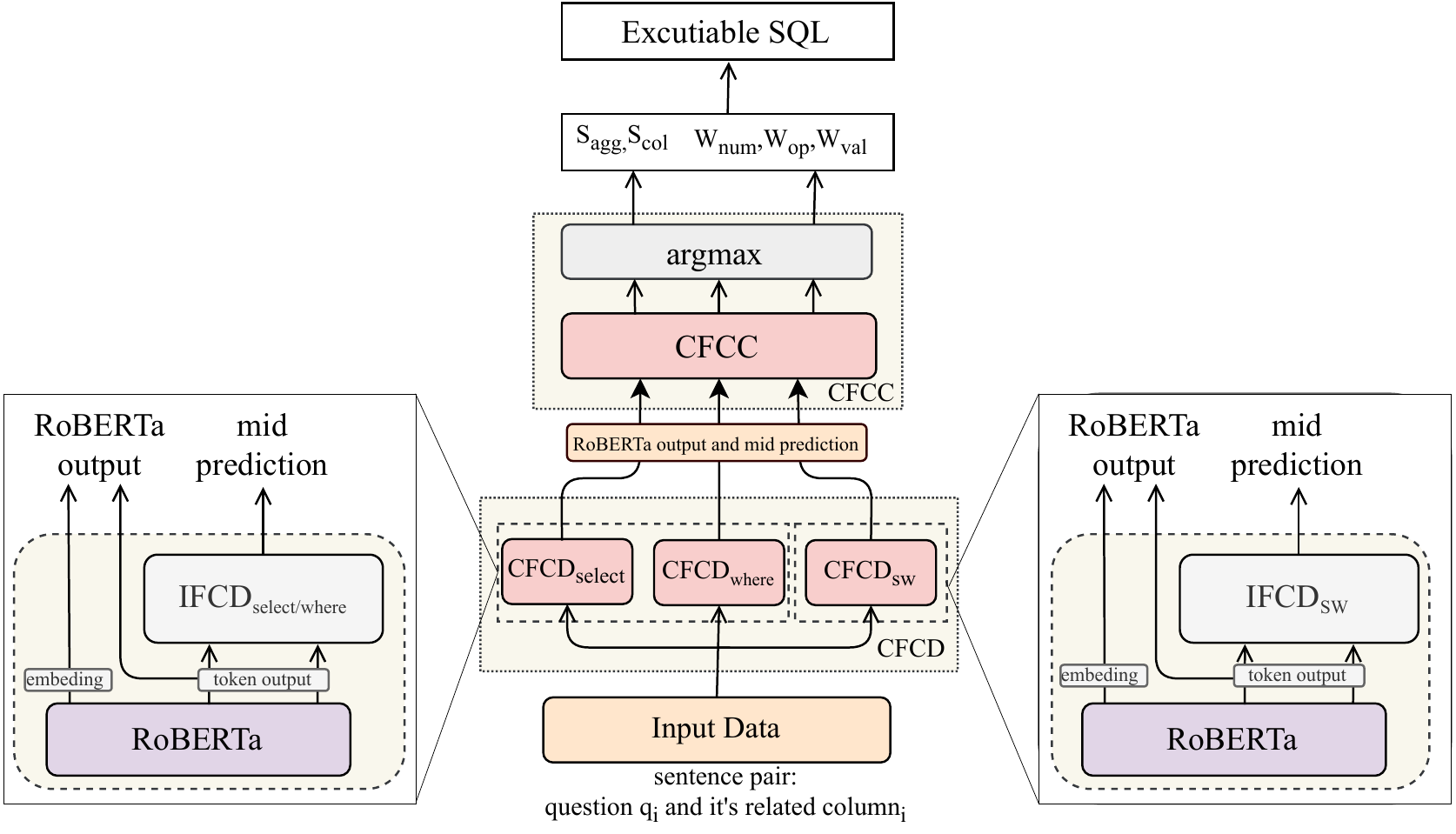}
\caption{The overall framework diagram of the CFCDC model shows the role of the three improved modules and the process of generating SQL} \label{fig2}
\end{figure*}

Then \textit{input} will be tokenized and encoded by language model such as BERT\cite{devlin2018bert} or RoBERTa\cite{liu1907roberta}. The form of the token sequence will be like this:
\begin{align}
[CLS],\textit{$x_1,...,x_n$},[SEP],\textit{$y_1,...,y_n$},[SEP]
\end{align}
where $x_i$ denotes $contact((c_i,\gamma_i,\delta_i),q_i)$ and $y_i$ denotes the tokens of question \textit{q}. This token sequence is the final input to our model. The overall structure of our model is shown in Fig.\ref{fig2}.
\subsection{Clause Feature Correlation Decoupling}
We present a framework called the Clause Feature Correlation Decouple (CFCD) to address the weak correlation between SELECT and WHERE clauses at the parameter level. The proposed CFCD module allocates specific language model for encoding and fine-tuning during the training phase. The main objective of the CFCD module is to divide the original multi-task system into three separate sub-modules, namely CFCD$_{select}$, CFCD$_{where}$, and CFCD$_{sw}$, which perform distinct functions. In particular, the CFCD$_{select}$ module is responsible for encoding and fine-tuning language models on columns in the SELECT clause ($S_c$), while the CFCD$_{where}$ module encodes and fine-tunes models on columns in the WHERE clause ($W_c$). The CFCD$_{sw}$ module, on the other hand, encodes and fine-tunes models on columns in the union of $W_c$ and $S_c$ ($R_c$).
\subsubsection{CFCD$_{select}$} 
\par The SELECT clause module serves the purpose of ranking clauses $c_i$, which form a part of the column set enlisted in the SELECT clause. The ranking mechanism is based on the number of occurrences of $c_i$ in the clauses of the SELECT clause, as per the criterion of q. During the classification procedure, the module determines the presence of SELECT clauses that include $c_i$ based on q, utilizing P($c_i\in S_c \mid q$) as the ranking score for sentence classification. Subsequent to obtaining the classification results from the linear layer, the CFCD$_{select}$ module leverages the following equation for predicting the number of select clauses (denoted by $n_s$):
\begin{align}
\hat{n}_s = \mathop{\arg\max}\limits_{n_s}P(n_s\mid q) = \mathop{\arg\max}\limits_{c_i \in S_c}P(n_s\mid c_i, q)P(c_i \in R_c \mid q) 
\end{align}
\par The CFCD model was trained using the Fast Gradient Method\cite{miyato2016adversarial}, which was found to improve the model's robustness. In order to optimize the performance of both the CFCD$_{select}$ and CFCD${where}$ components, a new loss function was employed. Specifically, the language model's output after multiple dropout, where $Y$ was used as the label, while $b_1$ and $b_2$ were the outputs of task $t_{i,1}$ and $t_{i,2}$, respectively. $f_c(x)$ was defined as a combination of the CrossEntropy function, and the KL divergence between two vectors, denoted by $KL_{div}$. Hyper-parameters $\lambda$ and $\mu$ were also introduced to tune the performance of the model. The resulting loss function can be expressed as follows:
\begin{align}
&loss_1 = 0.5*(f_c(t_{i,1},Y)+f_c(t_{i,2},Y))\\
&loss_2 = \lambda*\sum_i KL_{div}(t_{i,1},t_{i,2})\\
&loss_3 = \mu*KL_{div}(b_1,b_2)\\
&loss = loss_1+loss_2+loss_3
\end{align}
\par The deactivation of network parameters through the use of dropout has been observed to entail the randomization of loss functions in each gradient update, with function (4) representing the empirical loss, and functions (5) and (6) denoting the random losses. Through gradual optimization, the model is able to retain a majority of its optimal parameters, consequently improving accuracy and averting over-fitting during the training process. Notably, the loss function bears similarities to R-drop, as detailed in Wu's work\cite{wu2021r}.
\subsubsection{CFCD$_{where}$}
This section presents a module that assesses the ranking of clauses $c_i$ that are part of a set of WHERE clauses. The ranking is based on the frequency of occurrence of $c_i$ in the WHERE clauses of a query $q$. To determine whether WHERE clauses contain $c_i$ in the classification process, the module employs P($c_i\in W_c \mid q$). The training procedure and calculation equation for CFCD$_{where}$ closely resemble those for $CFCD_{select}$.
\subsubsection{CFCD$_{sw}$}
The objective of this module is to establish a ranking for clauses $c_i$ that are components of the SELECT clause and WHERE clause in a given query. This ranking is established by considering the number of occurrences of $c_i$ in both the SELECT and WHERE clauses based on a specific query, denoted by $q$. The ranking score is determined by the conditional probability of $c_i$ being present in the SW set, given the specific question $q$, denoted by P($c_i\in R_c \mid q$). Upon obtaining the classification of the linear layer, the module employs the following equation to predict the number of SELECT and WHERE clauses, denoted by $n_{sw}$.
\begin{align}
\hat{n}_{sw} = \mathop{\arg\max}\limits_{n_{sw}}P(n_{sw}\mid q)= \mathop{\arg\max}\limits_{c_i \in R_c}P(n_s\mid c_i, q)P(c_i \in R_c \mid q)
\end{align}
\par By utilizing CFCD modules, it becomes feasible to convert the original multi-task learning paradigm into three comparatively autonomous modules catering to individual functions. Integration of CFCD module not only resolves the issue of weak correlation in the original model but also enhances the efficiency of multi-task learning.
\begin{figure*}
\centering
\includegraphics[width=0.8\textwidth]{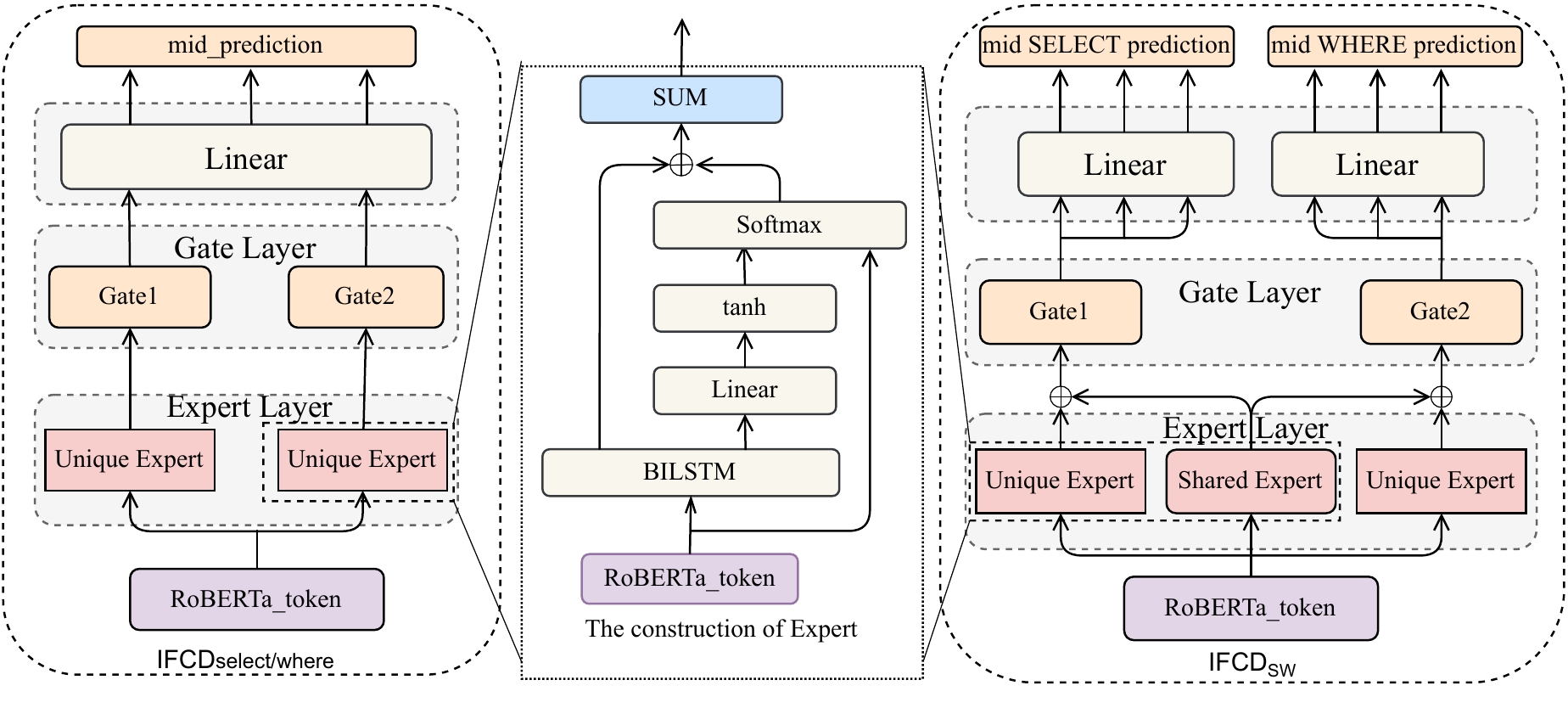}
\label{fig.3}
\caption{The structure of Implicit Feature Correlation Decoupling and its experts.} 
\end{figure*}
\subsection{Implicit Feature Correlation Decoupling}
NL2SQL models comprise not only explicit correlations among clauses but also implicit correlations among sub-tasks within a specific clause. The current implementation of HydraNet does not account for the implicit correlations between related sub-tasks. To address this issue, we propose the use of the Implicit Feature Correlation Decoupling (IFCD) technique in the CFCD module. By incorporating IFCD, we aim to address the lack of consideration of implicit correlations and improve the overall performance of HydraNet.

\par The presented research describes IFCD, a multi-task learning network whose structure is depicted in Fig.3 (\ref{fig.3}), and is based on Progressive Layered Extraction\cite{tang2020progressive}. The network architecture of IFCD comprises a shared expert at the bottom and various experts and gate structures for specific tasks at the top. The shared expert is utilized in different sub-tasks to extract the output of the language model, thereby enabling the identification of the overall pattern. For each sub-task, unique experts capture the semantic features of that sub-task, and gates are used to combine the output of the shared expert with the output of the respective unique expert. The resulting combination can replace the first token sequence of the language model for classification and, thus, enable the extraction of implicit correlation features.
\par In order to obtain further semantic information about the problem and the table clause, we use attention-based Bi-LSTM\cite{zhou2016attention} for the initial processing of the last layer of the BERT output token $b_i$.
\begin{align}
(z_1,z_2...,z_n)= Bi-LSTM (b_1,b_2...,b_n)
\end{align}
\par Following the output of the Bi-Directional Long Short-Term Memory (Bi-LSTM) $z_i$, the attention mechanism is employed to process the outputs of both the Bi-LSTM and Bidirectional Encoder Representations from Transformers (BERT) models. Specifically, the output of an expert, denoted as $h_c$, is incorporated in the attention calculation, where $W_i$ and $\theta$ represent the weight matrix and bias, respectively. Further, $mask_i$ represents the output of BERT's segment embedding, while $c$ and $a_i$ are hyper-parameters and weight vectors, respectively. The expert's query and key vectors are represented as $query_i$ and $key_i$, respectively. The length of the output vector is represented by the variable $n$.
\begin{align}
&query_i = W_i^T z_i+\theta \\
&key_i = Tanh(query_i) \\
a_i = softmax( &query_i*key_i*mask_i - c*(1-mask_i)) \\
&\quad h_c = \sum_{i=1}^{n} a_i
\end{align}
\par Then, the gating module $g_i$ is used to combine the shared expert $e_s$ and the sub-task unique expert $e_t$ of the sub-tasks and calculate their weighted results.
\begin{align}
out = g_{i,0}*e_s+g_{i,1}*e_t
\end{align}
\par In the optimization of IFCD, we propose a new loss function, which consists of two parts. For the loss of each sub-task $l_i$, the loss function is expressed as follows:
\begin{align}
\label{loss}
loss = 0.5*\sum_i l_i+0.5*awl(l_i)
\end{align}
$awl$ is an auto weight loss function for multi-task learning\cite{liebel2018auxiliary}. As a result of using IFCD, we are able to solve the implicit correlation extraction problem of related sub-tasks in the previous slot-filling method.
\subsection{Clause Feature Correlation Coupling}
\begin{figure}
\centering
\label{fig4}
\includegraphics[width=0.75\textwidth]{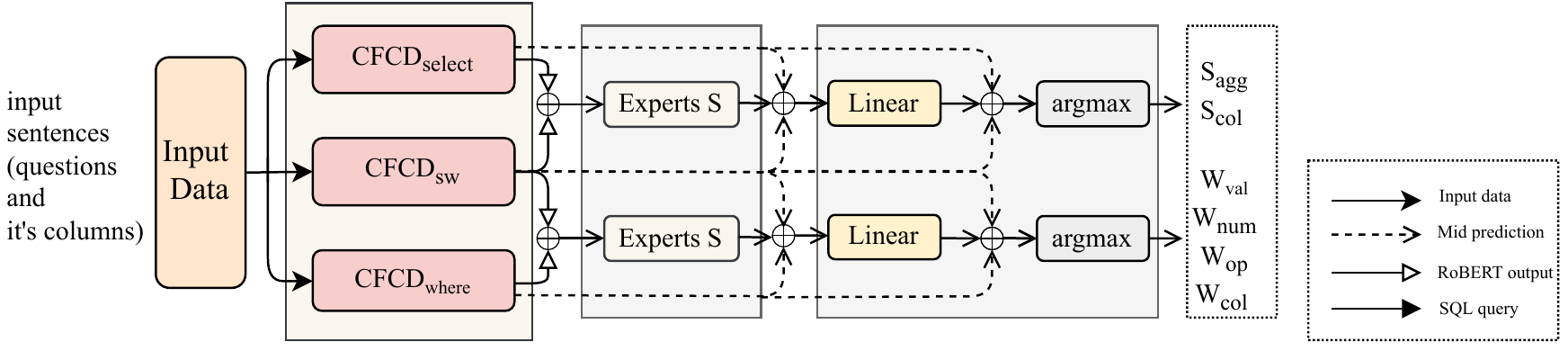}
\caption{The construct of clause correlation Coupling module, which is responsible for coupling different CFCD modules and output executable SQL query} 
\end{figure}
The addition of CFCD to HydraNet addresses the issue of low correlation and negative migration among explicit clauses, while IFCD resolves the problem of neglecting implicit feature correlations between component predictions. However, the output from these two modules cannot be transformed into executable SQL statements. To integrate the output from various CFCD modules, we employ the Clause Feature Correlation Coupling (CFCC) module to generate SQL statements based on the output of the aforementioned modules.
The CFCC architecture is presented in Fig 4(\ref{fig4}). The encoder of CFCC comprises three trained CFCC modules with its IFCD. During the prediction process, CFCC utilizes the output of CFCD modules to make classifications and predictions for different components. The first step in the process involves importing the CFCD modules and then processing the output of each imported CFCD module. The encoding outputs of CFCD$_{select}$ and CFCD$_{sw}$ module are utilized to form a new vector OS, while the encoding outputs of CFCD$_{where}$ and CFCD$_{sw}$ module are used to create another vector OW. These vectors serve as inputs for the extraction of the subsequent coupling mode.
CFCC is a neural network that couples the outputs of CFCD$_{select}$, CFCD$_{where}$, and CFCD$_{sw}$. CFCC employs two multi-task shared experts to extract the features of SELECT clause and WHERE clause. The expert dedicated to the sub-task extracts specific task features for the classification tasks of different components. The calculation equation of the expert is similar to IFCD.

After the expert couples the output of the CFCD module, it performs the prediction task of SQL statements. For the prediction of SQL statements, we use the expert's output and the CFCD's output to conduct weighted voting. As part of the SQL prediction process, the following steps are taken:

\begin{itemize}
\item Calculate the WHERE NUM and SELECT NUM using the coupling output of the expert and the weighted output of CFCD.
\item Based on the SELECT NUM, rank the $n_s$ columns, select the $N_S$ columns with higher scores, and set them as ($AGG_1$, $SC_1$), ($AGG_2$, $SC_2$), ..., ($AGG_n$, $SC_n$), where $AGG$ represents $SELECT_{agg}$ and $SC$ represents $SELECT_{col}$.
\item Based on the WHERE NUM, sort the WHERE column and select the top $n_m$ columns. The WHERE clause is set as ($WC_1$, $OP_1$, $VAL_1$), ($WC_2$, $OP_2$, $VAL_2$), ..., ($WC_n$, $OP_n$, $VAL_n$), where $OP$ represents $WHERE_{op}$, $WC$ represents $WHERE_{col}$, and $VAL$ represents $WHERE_{val}$.
\item Use T = ($t_1$, $t_2$, ..., $t_n$) to represent the union of the selected SELECT clause and WHERE clause, and use T to generate SQL statements.
\end{itemize}

\section{Experiment}
\subsection{Dataset}
The WikiSQL dataset, as described in \cite{DBLP:journals/corr/abs-1709-00103}, is a sizable collection of semantic parsing instances comprising 56,335 unique natural language to structured query language (NL2SQL) groups for training, 8,421 for validation, and 15,878 for testing. This dataset is evaluated based on two metrics: logical accuracy (LF) and execution accuracy (EX), which are akin to those employed in HydraNet\cite{lyu2020hybrid}.

\subsection{Implementation Details}
We use Pytorch to implement the overall model architecture. The word embedding of our sub-system model is randomly initialized by the RoBERTa\cite{liu1907roberta} model and fine-tuned during training. The batch size of the CFCD and the total CFCC module is 256. we choose the Adam\cite{kingma2014adam} optimizer to optimize our model and set the learning rate as 3.0e-5, the dropout rate is 0.2. We use 8X RTX3090 NVIDIA GPUs to train our CFCD modules, which costs 26 hours total (CFCD$_{select}$ costs 7 hours, CFCD$_{where}$costs 9 hours, CFCD$_{sw}$costs 10 hours). Furthermore, 4X RTX3090 NVIDIA GPUs are used to train the CFCC module, which costs 51 hours.

\subsection{Comparison with Existing Methods}
In this study, we conducted a comparative analysis of the CFCDC model with several existing models of WikiSQL:(1)Seq2SQL\cite{DBLP:journals/corr/abs-1709-00103} is the baseline for the WikiSQL dataset; (2)SQLNet\cite{DBLP:journals/corr/abs-1711-04436} is a sketch-based model; (3)TypeSQL\cite{yu2018typesql}Use the type information extracted from the text to better understand the entities and numbers in the problem; (4)RAT-SQL\cite{wang2019rat} uses relationship aware self-attention. (5)SQLova\cite{hwang2019comprehensive} first uses a pre-trained-language model to encode and integrate it into a sketch-based method; (6)X-SQL\cite{he2019x} uses contextual embedding in order to enhance the structural clause representation; (7)HydraNet\cite{lyu2020hybrid} uses column-wise ranking and decoding and is the baseline of our model; (8)IESQL\cite{hui2021improving} proposes a method of extracting information from NL2SQL, and solves this task by extracting relations based on sequence tags.
\begin{table}[h!]
\centering
    \label{tab:table1}
    \begin{tabular}{c | c c|  c c c c c}
      \hline
       & Dev & Test & &&Test\\
       \textbf{Model} &\textbf{LF} \quad \textbf{EX} &\textbf{LF}\quad\textbf{EX} &$S_{col}$
       &$S_{agg}$
       &$W_{col}$ 
       &$W_{op}$ &$W_{val}$\\
      \hline
      Seq2SQL &49.5\quad 60.8&48.3\quad59.4 &-&-&-&-&-\\
      SQLNet &63.2\quad69.8& 61.3\quad68.0 &-&-&-&-&-\\
      TypeSQL &68.0\quad74.5&66.7\quad 73.5&-&-&-&-&-\\
      RAT-SQL &73.6\quad82.0 &75.4\quad81.4&-&-&-&-&-\\
      SQLova  &81.6\quad87.2 &80.7\quad86.2 &96.8&90.6&94.3&97.3&95.4\\
      X-SQL &83.8\quad89.5 & 83.3\quad88.7 &97.2&91.1& 95.4&97.6&96.6\\
      HydraNet &83.6\quad89.1&83.8\quad89.2&97.6&91.4&97.2&97.5&97.6\\
      IESQL &84.6\quad89.7 &\textbf{84.6}\quad88.8&97.6&90.7&\textbf{96.4}&\textbf{97.7}&\textbf{96.7}\\
      CFCDC &\textbf{84.7}\quad \textbf{90.0} &84.5\quad\textbf{89.6}&\textbf{97.9}&\textbf{91.6}&95.7&97.6&96.5\\
      
      \hline 
      SQLova+EG &84.2\quad90.2 &83.6\quad89.6&96.5&90.4&95.5&95.8&95.9\\
      IESQL+EG &85.8\quad91.6 &85.6\quad91.2&97.6&90.7&\textbf{97.9}&\textbf{98.5}&\textbf{98.3}\\
      X-SQL+EG &86.2\quad92.3 & 86.0\quad91.8&97.2&91.1&97.2&97.5&97.9\\
      HydraNet+EG &86.6\quad92.4&86.5\quad92.2&97.6&91.4&97.2&97.5&97.6\\
      CFCDC+EG &\textbf{87.2}\quad\textbf{92.8}&\textbf{87.1}\quad\textbf{92.5} &\textbf{97.9}&\textbf{91.6}&97.5&97.6&97.7\\
      \hline
    \end{tabular}
 \caption{Logic form (LF), Execution (EX) accuracy and WikiSQL components accuracy of our model and other competitors in the validation set and test set. Best results in bold. EG: execution-guided decoding. '-' means no data is given in the model's original paper}
\end{table}
Table 1 provides a comprehensive overview of the accuracy comparison between our model and other competitors in the dataset. Initially, our model's accuracy was compared to other advanced models on the WikiSQL benchmark, without the utilization of execution-guided (EG) decoding on the test set. Our model's accuracy, as displayed in Table 1 (\ref{tab:table1}), outperformed most state-of-the-art natural language to structured query language (NL2SQL) models. When EG decoding was incorporated, our model outperformed all competitors in all indicators, demonstrating the impact of hard decouple correlation and couple correlation in the NL2SQL field. Furthermore, we compared the accuracy of several models for various SQL components, with and without EG decoding. As indicated in Table 1 (\ref{tab:table1}), our model exhibited superior performance for S$_{col}$ components, while for other components, our model was second only to IESQL with the guidance of EG.

\subsection{Ablation Studies}
Table 2 presents a comprehensive analysis of the impact of individual components in the model on the overall task performance.
\begin{table}[h!]
  \centering
    \label{tab:table2}
    \begin{tabular}{c | c c| c c c c c}
      \hline
       & Dev & Test &&&Test\\
       \textbf{Model}&\textbf{LF} \quad \textbf{EX} &\textbf{LF}\quad\textbf{EX}&$S_{col}$
       &$S_{agg}$
       &$W_{col}$ 
       &$W_{op}$ &$W_{val}$\\
      \hline
      HydraNet  &83.6\quad 89.1&83.8\quad89.2 &97.6&91.4&95.3&97.4&96.1\\
      (+)IFCD &84.0\quad89.4& 83.7\quad88.8  &97.5 & 91.3&95.0&97.4&96.0\\
      (+)CFCD &84.3\quad89.5&84.2\quad\textbf{89.7} &97.8&91.4&95.4&97.4&96.3\\
      (+)CFCD + IFCD  &82.1\quad86.8 &81.2\quad86.2 &97.7&91.4&95.4&97.4&96.2\\
      (+)CFCD + CFCC &84.4\quad\textbf{90.0} &84.3\quad89.6 &97.8&91.4&95.4&97.4&96.3\\
      (+)CFCD + IFCD + CFCC&\textbf{84.7}\quad\textbf{90.0} &\textbf{84.5}\quad89.6&\textbf{97.9}&\textbf{91.6}&\textbf{95.7}&\textbf{97.5}&\textbf{96.5}\\
      \hline 
      baseline(HydraNet)+EG &86.6\quad 92.4&86.5\quad92.2 &97.6&91.4&97.2&97.5&97.6\\
      (+)IFCD + EG&86.9\quad92.4& 86.7\quad92.2 &97.5 & 91.3&97.2&97.5&97.5\\
      (+)CFCD + EG &86.9\quad 92.6&86.8\quad92.4 &97.8&91.4&97.4&97.5&\textbf{97.7}\\
      (+)CFCD + IFCD + EG &86.8\quad92.3 &86.6\quad92.2
      &97.7&91.1&97.2&97.4&97.6\\
      (+)CFCD + CFCC + EG &86.9\quad\textbf{92.8} &86.8\quad\textbf{92.5}&97.8&91.4&97.4&97.5&\textbf{97.7}\\
      (+)CFCD + IFCD + CFCC + EG&\textbf{87.2}\quad\textbf{92.8} &\textbf{87.1}\quad\textbf{92.5}&\textbf{97.9}&\textbf{91.6}&\textbf{97.5}&\textbf{97.6}&\textbf{97.7}\\
      \hline 
    \end{tabular}
 \caption{Logic form (LF), Execution (EX) accuracy and WikiSQL components accuracy of baseline with different components in the validation set and test set. (+)CFCC + IFCD + CFCC is our model, Best results are in bold. EG: execution-guided decoding.}
\end{table}
The results demonstrate that the CFCC module significantly improves the accuracy of all tasks, particularly in predicting WHERE components such as $W_{col}$, $W_{op}$ and $W_{val}$, while the IFCD and CFCC modules show promising results. An ablation study was conducted by introducing three new modules, namely the CFCD, IFCD, and CFCC modules, to the baseline model (HydraNet). The results indicate that the addition of IFCD may negatively affect the overall accuracy of the model execution, but using CFCC modules in combination with CFCD modules can significantly enhance the model's overall accuracy. Furthermore, IFCD can be employed to significantly enhance the logical accuracy of the validation and test sets in execution-guided decoding cases.

\subsection{Impact of Simplifying Method}
In our study, we investigated the impact of Implicit Feature Correlation Decoupling  (IFCD) experts on execution-guided (EG) decoding accuracy. Our research aimed to simplify the expertise of the different sub-systems in IFCD. 
\begin{table}
  \centering
    \label{tab:table3}
    \begin{tabular}{c | c c| c c c c c}
      \hline
       & Dev & Test &&&Test\\
       \textbf{Model}&\textbf{LF} \quad \textbf{EX} &\textbf{LF}\quad\textbf{EX}&$S_{col}$
       &$S_{agg}$
       &$W_{col}$ 
       &$W_{op}$ &$W_{val}$\\
      \hline
      CFCDC$_{simplified}^S$ &84.3\quad 89.6&84.0\quad89.3 &97.5&91.3&95.3&97.4&96.1\\
      CFCDC$_{simplified}^W$ &84.3\quad89.6& 83.7\quad89.2&97.6 & 91.4&95.0&97.4&96.0\\
      CFCDC &\textbf{84.7}\quad\textbf{90.0} &\textbf{84.5}\quad\textbf{89.6}&\textbf{97.9}&\textbf{91.6}&\textbf{95.7}&\textbf{97.5}&\textbf{96.5}\\
      \hline
      CFCDC$_{simplified}^S$ + EG  &87.0\quad\textbf{92.8} &86.6\quad92.3 &97.5&91.3&97.2&97.5&97.6\\
      CFCDC$_{simplified}^W$ + EG &86.8\quad92.4 &86.8\quad92.3 &97.6& 91.4&97.2&97.5&97.5\\
      CFCDC+EG &\textbf{87.2}\quad\textbf{92.8} &\textbf{87.1}\quad\textbf{92.5}&\textbf{97.9}&\textbf{91.6}&\textbf{97.5}&\textbf{97.6}&\textbf{97.7}\\
      \hline
    \end{tabular}
 \caption{Logic form (LF), Execution (EX) accuracy and WikiSQL components accuracy of our final models and models that simplify the experts in IFCD in the validation set and test set. Best results in bold. CFCDC is the final model. EG: Execution-guided decoding.}
\end{table}
The results presented in Table 3(\ref{tab:table3}), indicate that simplifying the SELECT part of IFCD had only a minor effect on its final execution accuracy. Conversely, simplifying either the SELECT or WHERE part of IFCD led to varying degrees of decline in logic accuracy, albeit small differences when EG was present. We hypothesize that the expert's input in IFCD may not be closely linked to the EG scenario, thus limiting its impact. 
\section{Conclusion}
This paper presents an investigation of the application of CFCD, IFCD, and CFCC to enhance the effectiveness of the NL2SQL task. To address the fundamental phenomena encountered in the NL2SQL task, including the weak correlation between SELECT and WHERE clauses and the seesaw phenomenon resulting from implicit features correlation between sub-tasks in the SELECT (WHERE) clause, we propose a network structure named CFCDC. The SELECT and WHERE clauses are trained independently to achieve hard feature representation decoupling at the parameter level, whereas a composite expert network is employed to achieve soft feature representation decoupling and capture the implicit correlation between sub-tasks, thereby mitigating the seesaw phenomenon associated with using a multi-task learning approach. Our model outperforms the original model by 1.1\% (validation set) and 0.7\% (test set), indicating the potential for improvement in slot-filling models in the NL2SQL domain. We anticipate that this study will provide insights into how to handle the correlation dependencies of different tasks in NL2SQL, ultimately enhancing the overall model performance.

\bibliographystyle{splncs04} 
\bibliography{conference} 

\begin{thebibliography}{10}
\providecommand{\url}[1]{\texttt{#1}}
\providecommand{\urlprefix}{URL }
\providecommand{\doi}[1]{https://doi.org/#1}

\bibitem{devlin2018bert}
Devlin, J., Chang, M.W., Lee, K., Toutanova, K.: Bert: Pre-training of deep
  bidirectional transformers for language understanding. arXiv preprint
  arXiv:1810.04805  (2018)

\bibitem{dong2018coarse}
Dong, L., Lapata, M.: Coarse-to-fine decoding for neural semantic parsing. In:
  56th Annual Meeting of the Association for Computational Linguistics. pp.
  731--742. Association for Computational Linguistics (2018)

\bibitem{he2019x}
He, P., Mao, Y., Chakrabarti, K., Chen, W.: X-sql: reinforce schema
  representation with context. arXiv preprint arXiv:1908.08113  (2019)

\bibitem{hui2021improving}
Hui, B., Shi, X., Geng, R., Li, B., Li, Y., Sun, J., Zhu, X.: Improving
  text-to-sql with schema dependency learning. arXiv preprint arXiv:2103.04399
  (2021)

\bibitem{hwang2019comprehensive}
Hwang, W., Yim, J., Park, S., Seo, M.: A comprehensive exploration on wikisql
  with table-aware word contextualization. arXiv preprint arXiv:1902.01069
  (2019)

\bibitem{kingma2014adam}
Kingma, D.P., Ba, J.: Adam: A method for stochastic optimization. arXiv
  preprint arXiv:1412.6980  (2014)

\bibitem{liebel2018auxiliary}
Liebel, L., K{\"o}rner, M.: Auxiliary tasks in multi-task learning. arXiv
  preprint arXiv:1805.06334  (2018)

\bibitem{liu1907roberta}
Liu, Y., Ott, M., Goyal, N., Du, J., Joshi, M., Chen, D., Levy, O., Lewis, M.,
  Zettlemoyer, L., Stoyanov, V.: Roberta: A robustly optimized bert pretraining
  approach. arxiv 2019. arXiv preprint arXiv:1907.11692  (1907)

\bibitem{lyu2020hybrid}
Lyu, Q., Chakrabarti, K., Hathi, S., Kundu, S., Zhang, J., Chen, Z.: Hybrid
  ranking network for text-to-sql. arXiv preprint arXiv:2008.04759  (2020)

\bibitem{miyato2016adversarial}
Miyato, T., Dai, A.M., Goodfellow, I.: Adversarial training methods for
  semi-supervised text classification. arXiv preprint arXiv:1605.07725  (2016)

\bibitem{sun2018semantic}
Sun, Y., Tang, D., Duan, N., Ji, J., Cao, G., Feng, X., Qin, B., Liu, T., Zhou,
  M.: Semantic parsing with syntax-and table-aware sql generation. arXiv
  preprint arXiv:1804.08338  (2018)

\bibitem{tang2020progressive}
Tang, H., Liu, J., Zhao, M., Gong, X.: Progressive layered extraction (ple): A
  novel multi-task learning (mtl) model for personalized recommendations. In:
  Fourteenth ACM Conference on Recommender Systems. pp. 269--278 (2020)

\bibitem{torrey2010transfer}
Torrey, L., Shavlik, J.: Transfer learning. In: Handbook of research on machine
  learning applications and trends: algorithms, methods, and techniques, pp.
  242--264. IGI global (2010)

\bibitem{wang2019rat}
Wang, B., Shin, R., Liu, X., Polozov, O., Richardson, M.: Rat-sql:
  Relation-aware schema encoding and linking for text-to-sql parsers. arXiv
  preprint arXiv:1911.04942  (2019)

\bibitem{wu2021r}
Wu, L., Li, J., Wang, Y., Meng, Q., Qin, T., Chen, W., Zhang, M., Liu, T.Y.,
  et~al.: R-drop: Regularized dropout for neural networks. Advances in Neural
  Information Processing Systems  \textbf{34},  10890--10905 (2021)

\bibitem{DBLP:journals/corr/abs-1711-04436}
Xu, X., Liu, C., Song, D.: Sqlnet: Generating structured queries from natural
  language without reinforcement learning. CoRR  \textbf{abs/1711.04436}
  (2017), \url{http://arxiv.org/abs/1711.04436}

\bibitem{yu2018typesql}
Yu, T., Li, Z., Zhang, Z., Zhang, R., Radev, D.: Typesql: Knowledge-based
  type-aware neural text-to-sql generation. arXiv preprint arXiv:1804.09769
  (2018)

\bibitem{DBLP:journals/corr/abs-1709-00103}
Zhong, V., Xiong, C., Socher, R.: Seq2sql: Generating structured queries from
  natural language using reinforcement learning. CoRR  \textbf{abs/1709.00103}
  (2017), \url{http://arxiv.org/abs/1709.00103}

\bibitem{zhou2016attention}
Zhou, P., Shi, W., Tian, J., Qi, Z., Li, B., Hao, H., Xu, B.: Attention-based
  bidirectional long short-term memory networks for relation classification.
  In: Proceedings of the 54th annual meeting of the association for
  computational linguistics (volume 2: Short papers). pp. 207--212 (2016)

\end{thebibliography}
\end{document}